\documentclass[10pt, a4paper]{article}
\usepackage{lrec-coling2024} 

\usepackage{multirow}
\usepackage{booktabs}
\usepackage{amsmath}
\usepackage{amssymb}
\usepackage{graphicx}
\usepackage{subfig}
\usepackage{bm}

\usepackage{tikz}
\usepackage{pgfplots}
\usepackage{booktabs}
\usepackage{tabularx}
\usepackage{subcaption}

\usepackage{enumitem}
\setenumerate[1]{itemsep=0pt,partopsep=0pt,parsep=\parskip,topsep=5pt}
\setitemize[1]{itemsep=0pt,partopsep=0pt,parsep=\parskip,topsep=5pt}
\setdescription{itemsep=0pt,partopsep=0pt,parsep=\parskip,topsep=5pt}

\title{How to Understand ``Support''? An Implicit-enhanced Causal Inference Approach for Weakly-supervised Phrase Grounding}

\name{Jiamin Luo$^*$, Jianing Zhao$^*$\thanks{\@ \@  $^*$Equal Contribution.}, Jingjing Wang$^\dagger$\thanks{\@ \@  $^\dagger$Corresponding Author: Jingjing Wang.}, Guodong Zhou} 

\address{School of Computer Science and Technology, Soochow University, China \\
         No.1, Shizi Street, Suzhou City, Jiangsu Province, China \\
         \{20204027003, jnzhao1106\}@stu.suda.edu.cn, \{djingwang, gdzhou\}@suda.edu.cn\\}

\abstract{
Weakly-supervised Phrase Grounding (WPG) is an emerging task of inferring the fine-grained phrase-region matching, while merely leveraging the coarse-grained sentence-image pairs for training. However, existing studies on WPG largely ignore the implicit phrase-region matching relations, which are crucial for evaluating the capability of models in understanding the deep multimodal semantics. To this end, this paper proposes an \textbf{I}mplicit-\textbf{E}nhanced \textbf{C}ausal \textbf{I}nference (IECI) approach to address the challenges of modeling the implicit relations and highlighting them beyond the explicit. Specifically, this approach leverages both the intervention and counterfactual techniques to tackle the above two challenges respectively. Furthermore, a high-quality implicit-enhanced dataset is annotated to evaluate IECI and detailed evaluations show the great advantages of IECI over the state-of-the-art baselines. Particularly, we observe an interesting finding that IECI outperforms the advanced multimodal LLMs by a large margin on this implicit-enhanced dataset, which may facilitate more research to evaluate the multimodal LLMs in this direction.
 \\ \newline \Keywords{Weakly-supervised Phrase Grounding, Implicit Phrase-Region Matching, Causal Inference}}

\begin{document}

\maketitleabstract

\section{Introduction} 
\label{sec:intro}
Phrase Grounding (PG)~\cite{WangLHL19}, a fundamental task in the field of multimodal learning, aims to find all the regions within an image that correspond to various phrases present in a given sentence. This correspondence serves as a fundamental foundation for numerous vision-language tasks, including image captioning~\cite{Feng00L19a}, vision question answering~\cite{MunLSH18}, and visual dialog~\cite{GuoWZZW20}. However, PG heavily relies on expensive annotations of linking phrases to the corresponding image regions, which is labor-intensive and time-consuming. Thus, existing studies on PG mainly seek to address this task in the Weakly-supervised Phrase Grounding (WPG)~\cite{ChenGN18} setting which merely leverages coarse-grained sentence-image pairs during training while subsequently evaluates the performance on fine-grained phrase-region pairs, achieving substantial advancements.

\begin{figure}
\begin{center}
\subfloat{
\includegraphics[scale=0.60]{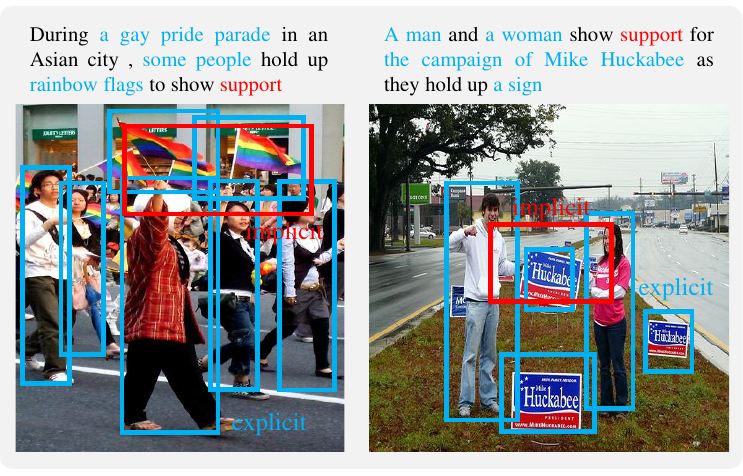}}
 \\
  \setlength{\belowcaptionskip}{-3.0 ex}
\caption{Two sentence-image pairs to illustrate the implicit (red phrases and boxes) and explicit (blue phrases and boxes) relations between phrases and regions.}
\label{fig:example}
\end{center}
\end{figure}

Despite this, these studies fail to deeply explore the semantic nature of phrases that some phrases often exhibit the implicit and intricate semantics, rendering it arduous for models to establish correct connections with image regions. Take Figure \ref{fig:example} as an example, the phrase ``\emph{support}'' necessitates the integration of the commonsense knowledge to precisely find its corresponding image region, i.e., red boxes. In this study, we refer to such correspondence as one type of implicit phrase-region matching relations\footnote{In this paper, this commonsense-involved implicit relation is named as ``\emph{commonsense understanding}''. Besides, we also propose another three implicit relations as shown in Figure \ref{fig:annotation}.}, which could be defined as phrases in the sentence that are not specific and explicit nominal phrases of the objects in the image, and the grasp of such implicit relations serves as a valuable evaluation of the model capability in understanding the deep multimodal semantics. Additionally, the weakly-supervised setting of WPG could bring the supervised noise problem~\cite{XiaoSL17} which may lead to more difficulties in capturing such implicit information, making it urgent to address the implicit relations problem in WPG. In this study, we contend that capturing such relations at least faces two main challenges.

For one thing, we argue that modeling the implicit relations is challenging. Still take the implicit relation sample in Figure \ref{fig:example} as the example, the implicit phrase ``\emph{support}'' actually only corresponds to a special small-scale red-box region (i.e., the people's arms with the held flags or signs), which is easily confounded by the other common regions and thus difficult to be predicted. Fortunately, in the literature, a few recent studies on other tasks (e.g., \citealp{WangHZS20a}) have also encountered with the similar confounding bias problem and suggested to leverage the causation-based approach for mitigating this confounding bias. Inspired by this, we believe that a well-behaved approach to WPG should take advantage of the causation-based approach (e.g., causal inference~\cite{pearl2018book}) to model the implicit relations.

For another, we argue that highlighting the implicit relations beyond the explicit is rather challenging. As exemplified in Figure \ref{fig:example}, it is obvious that the occurrence ratio of implicit relations (i.e., ``\emph{support}'') is significantly lower (about 1:9, see Section \ref{sec:datasets}) compared to explicit relations (i.e., ``\emph{some people}'', ``\emph{rainbow flags}'', ``\emph{a woman}'', ``\emph{a sign}''), which could mislead the model to prefer capturing the explicit phrase-region relations instead of the implicit. Therefore, we believe that a better-behaved causation-based approach to WPG should further consider this imbalance for better aligning the implicit phrase-region pairs.

To tackle the aforementioned challenges, this paper proposes a causation-based approach namely \textbf{I}mplicit-\textbf{E}nhanced \textbf{C}ausal \textbf{I}nference (IECI) for WPG. Specifically, this approach first leverages the intervention technique~\cite{pearl2018book} and proposes an implicit-aware deconfounded attention (IDA) block to model the implicit relations for mitigating the confounding bias. Furthermore, this approach leverages the counterfactual technique~\cite{pearl2018book} and proposes an implicit-aware counterfactual inference (ICI) block to highlight the implicit relations beyond the explicit for better aligning the implicit phrase-region pairs. Particularly, a high-quality implicit-enhanced dataset is annotated for benchmarking our IECI approach. The main contributions of our work are summarized as follows:
\begin{itemize}
    \item We are the first to address the implicit relations problem in WPG, and annotate a high-quality implicit-enhanced dataset to evaluate the ability of models in understanding deep multimodal semantics.  
    \item We propose a new IECI approach, which integrates both the intervention and counterfactual techniques for addressing the implicit challenges inside WPG. Detailed evaluations on our implicit-enhanced dataset demonstrate the superiority of our IECI approach over the state-of-the-art baselines.
    \item We observe an interesting finding that our IECI approach exhibits significant advantages compared to the advanced multimodal LLMs on the annotated implicit-enhanced dataset, which may further facilitate the evaluation of multimodal LLMs in this direction.
\end{itemize}

\section{Related Work}
\subsection{Phrase Grounding}
Phrase Grounding (PG) and Referring Expression Comprehension (REC)~\cite{KazemzadehOMB14} are two prevalent tasks within the field of Visual Grounding (VG)~\cite{RohrbachRHDS16}. While both tasks involve establishing relations between sentence phrases and image regions, PG focuses on predicting regions for all phrases within a sentence-image pair, whereas REC pertains to identifying a single region in the image corresponding to the given sentence. PG can be broadly categorized into two forms, i.e., one-stage models~\cite{YangXYLLH22,DengYCZL21} and two-stage models~\cite{ChenMXZC21,LiuWZH20}. Recognizing the expensive and difficult annotation of phrase-region, recent studies have predominantly shifted toward WPG. Early studies~\cite{ChenGN18,0006LZF18} primarily focus on directly learning the phrase-region relations. Besides, some studies~\cite{Liu0WZMH19,LiuLWZSH19,LiuWM021} acknowledge the importance of context cues, which exploit linguistic contexts to enforce cross-modal consistency. Moreover, \citet{DattaSRAPD19} introduce a ranking-loss to minimize the distances between associated sentence-image and maximize the distance between irrelevant pairs, inspiring numerous studies~\cite{GuptaVC0KH20,WangHLXY021,WangTSMY20,ChenZMM22} to employ contrastive learning techniques to predict phrase-region matching in the WPG task.

In summary, all the above studies always ignore the implicit phrase-region matching relations problem, which however holds significant potential for evaluating the ability of models in understanding the deep multimodal semantics.

\begin{figure*}
\begin{center}
\subfloat{
 \includegraphics[scale=0.47]{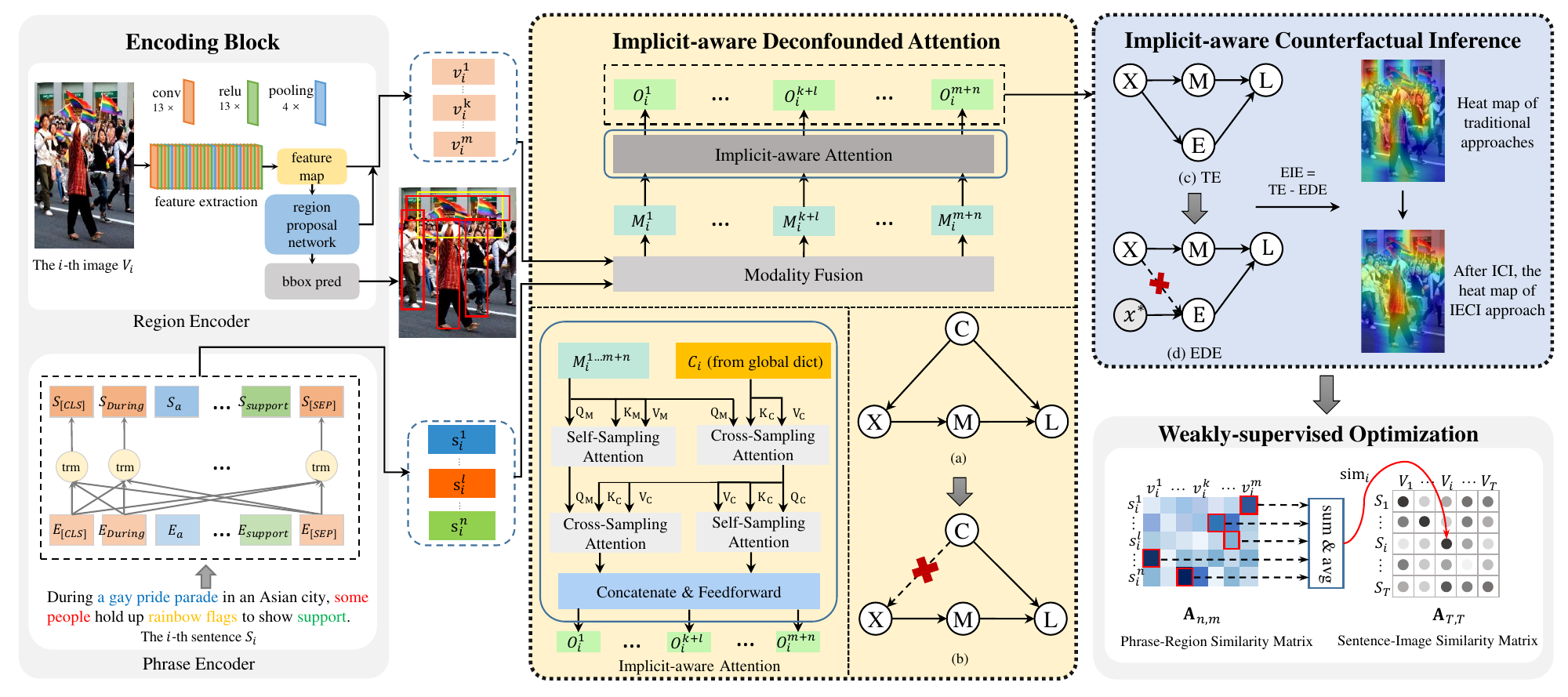}}
 \setlength{\belowcaptionskip}{-3.0 ex}
\caption{The overall framework of our proposed Implicit-Enhanced Causal Inference (IECI) approach. Wherein (a) and (b) are causal graphs for modeling the implicit relations (see Section \ref{sec:idca}), while (c) and (d) are those for highlighting the implicit relations beyond the explicit (see Section \ref{sec:icfi}).}
\label{fig:model}
\end{center}
\end{figure*} 

\subsection{Causal Inference}
In recent years, causal inference has sparked significant interest across a range of areas, including scene graph generation~\cite{TangNHSZ20}, semantic segmentation~\cite{ZhangZT0S20}, vision-language tasks~\cite{0016YXZPZ20}, etc. \citet{pearl2018book} have defined three levels of causality, encompassing intervention and counterfactual that are frequently employed to mitigate confounding bias and achieve unbiased estimations. For example, \citet{WangHZS20a} employ the causal intervention to deal with spurious correlation within datasets for visual common sense learning, and \citet{ZhangJWKZZYYW20} alleviate the spurious correlations between vision and language in vision-linguistic pre-training. For the strategy of intervention adjustment, \citet{YangZQ021} employ front-door adjustment to realize the causal intervention, while \citet{wang2021causal} use back-door adjustment to self-annotate the confounder in an unsupervised way. \citet{huang2022deconfounded} propose a confounder-agnostic approach to remove the confounding bias between language and location. In addition, \citet{TangHZ20} and \citet{NiuTZL0W21} employ counterfactual inference to alleviate long-tailed categories bias in image classification and language bias in VQA, respectively. 

Different from all above studies, our study is a pioneering effort in integrating both intervention and counterfactual techniques for multimodal tasks, where we leverage intervention to effectively mitigate the bias of confounding, and employ counterfactual to highlight the implicit matching relations.

\section{Implicit-Enhanced Causal Inference Approach}
In this section, we formulate the WPG task as follows. Given a collection of $T$ sentence-image pairs $(S,V)$, where $S=[S_1,...,S_T]$ and $V=[V_1,...,V_T]$. Each sentence $S_i$ consists of multiple phrases $S_i=[s_{i}^{1},..,s_{i}^{l},...,s_{i}^{n}]$, while each image $V_i$ comprises a set of regions $V_i=[v_{i}^{1},...,v_{i}^{k},...,v_{i}^{m}]$, where $n$ and $m$ represent the number of phrases and regions, respectively. The goal is to predict the region $v_{i}^{k}$ from the set of $m$ regions in image $V_i$ that corresponds to the given phrase $s_{i}^{l}$ in sentence $S_i$. However, under the WPG setting, we only have access to coarse-grained sentence-image pair $(S_i,V_i)$ for training, whereas fine-grained phrase-region pair $(s_{i}^{l},v_{i}^{k})$ is available during inference.

In this paper, we propose an \textbf{I}mplicit-\textbf{E}nhanced \textbf{C}ausal \textbf{I}nference (IECI) approach to model the implicit relations. Figure \ref{fig:model} shows the overall architecture of the proposed IECI approach, consisting of three major components: 1) Encoding Block, 2) Implicit-aware Deconfounded Attention (IDA) Block, 3) Implicit-aware Counterfactual Inference (ICI) Block. Prior to delving into the intricacies of the core components within IECI, we provide an overview of the encoding block.

\subsection{Encoding Block}
Given $T$ pairs of sentence and image, following the setting by \citet{GuptaVC0KH20}, BERT~\cite{DevlinCLT19} and Faster R-CNN~\cite{RenHGS15} are adopted to encode phrases and regions. 

\textbf{Phrase Encoder.} BERT-base model released by ~\citet{DevlinCLT19} is adopted as the phrase encoder, which is a light-weighting language encoding model. Specifically, the phrases are first extracted by following~\citet{PlummerWCCHL15}. Then, BERT is utilized to encode the sentence $S_i$, and finally the word vectors of all words in each phrase $s_{i}^{l}$ are averaged as the phrase encoding.

\textbf{Region Encoder.} Faster R-CNN is adopted as the region encoder. Specifically, Faster R-CNN first utilizes the convolutional network (i.e., ResNet~\cite{HeZRS16}) to compute a feature map of each image $V_i$. On this basis, a region proposal network is used to generate the encoding of the region $v_{i}^{k}$ along with the corresponding bounding box.

\subsection{Implicit-aware Deconfounded Attention Block}
\label{sec:idca}
In this study, we take advantage of the intervention technique~\cite{pearl2018book} and propose an \textbf{I}mplicit-aware \textbf{D}econfounded \textbf{A}ttention (IDA) block to model the implicit relations for mitigating the confounding bias inside WPG. Specifically, we address two crucial questions: 1) how to mitigate the confounding bias through the front-door adjustment strategy~\cite{pearl2018book}; 2) how to implement the front-door adjustment strategy in the WPG task. We will provide comprehensive answers to these two questions in the subsequent section, formulated as follows.

\textbf{Deconfounded Causal Graph} is leveraged to answer the question 1). As illustrated in Figure \ref{fig:model} (a), we formulate the causation among sentence-image pairs $\rm X$, multimodal knowledge $\rm M$, phrase-region locations $\rm L$, confounding factors $\rm C$. $\rm X \rightarrow \rm M \rightarrow \rm L$ denotes the desired causal effect from sentence-image pairs $\rm X$ to phrase-region locations $\rm L$, where multimodal knowledge $\rm M$ acts as a mediator. $\rm X \leftarrow \rm C \rightarrow \rm L$ denotes the causal effect from the invisible confounding factors $\rm C$ to sentence-image pairs $\rm X$ and phrase-region locations $\rm L$. 

We leverage \emph{do}-operator~\cite{pearl2018book} to mitigate the confounding bias $(\rm X,\rm C)$ present in the path $\rm M \rightarrow \rm L$. As shown in Figure \ref{fig:model} (b), we block the back-door path $\rm M \leftarrow \rm X \leftarrow \rm C \rightarrow \rm L$ under the condition of $\rm X$. Then, we leverage the front-door adjustment strategy to analyze the causal effect of $\rm X \rightarrow \rm L$, denoted as follows:
\begin{equation}
\label{equ:front_door}
\begin{aligned}
    &P({\rm L}=l|do({\rm X}=x))\\
    &=\sum_{m}P(m|x)\sum_{x}P(x)[P(l|x,m)]
\end{aligned}
\end{equation}

\begin{figure*}
\begin{center}
\subfloat{\includegraphics[scale=0.52]{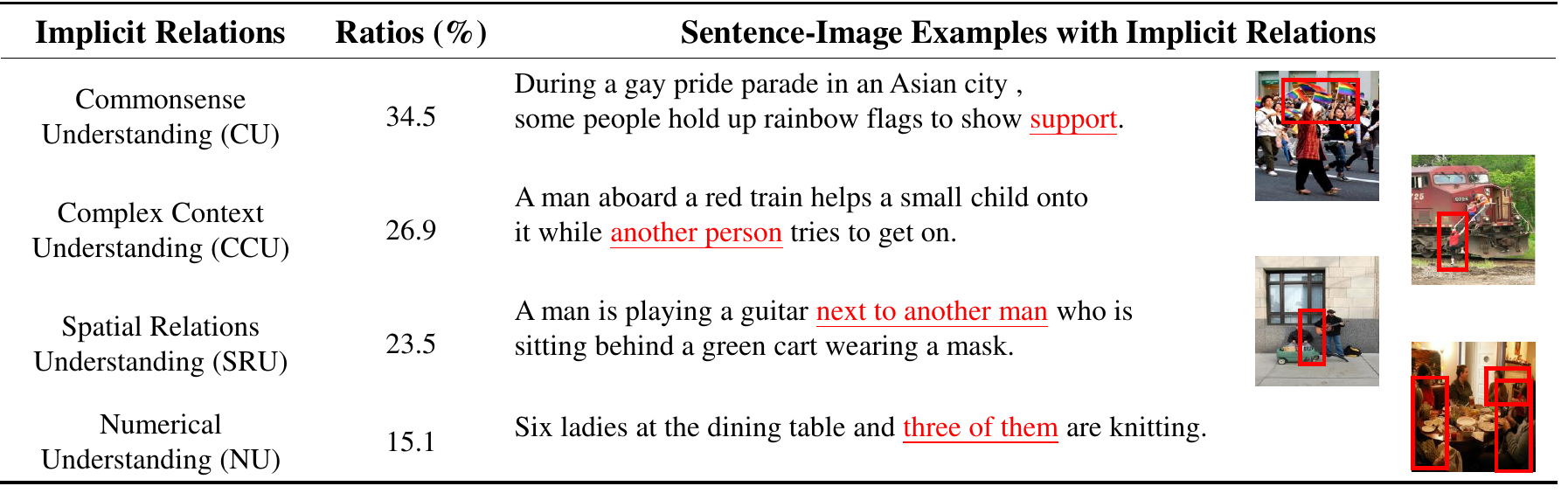}}
 \setlength{\belowcaptionskip}{-2.0 ex}
\caption{Four main types of the implicit phrase-region matching relations together with their corresponding ratios within the implicit phrase-region pairs.}
\label{fig:annotation}
\end{center}
\end{figure*} 

\textbf{Implicit-aware Attention} is leveraged to answer the question 2). On the basis of the front-door adjustment strategy in Eq.(\ref{equ:front_door}), we consider implementing it through the utilization of attention mechanisms. Considering the expensive computation of network forward propagation for all samples, we introduce the Normalized Weighted Geometric Mean (NWGM)~\cite{SrivastavaHKSS14,XuBKCCSZB15} approximation. Therefore, we can sample $\rm X$, $\rm M$ and complete $P({\rm L}|do({\rm X}))$ by feeding them into the network, and then leverage NWGM approximation to achieve the goal of Eq.(\ref{equ:front_door}), denoted as follows:
\begin{equation}
\label{equ:causal}
    P({\rm L}|do({\rm X})) \approx {\rm{softmax}}[g(\hat{\rm X}, \hat{\rm M})]
\end{equation}
where $g(.)$ is a network employed to parameterize the predictive distribution $P(l|x,m)$, which is followed by a softmax layer. Besides, $\hat{\rm M}=\sum_{m}P({\rm M}=m|h({\rm X})){\bm m}$ and $\hat{\rm X}=\sum_{x}P({\rm X}=x|f({\rm X})){\bm x}$ represent the estimations of self-sampling and cross-sampling, respectively. The variables $m,x$ correspond to the embedding vectors of $\bm m,\bm x$. The query embedding functions $h(.)$ and $f(.)$ are utilized to transform the input $X$ into two distinct query sets, which can be parameterized as networks. Consequently, we leverage attention mechanisms to estimate the self-sampling $\hat{\rm M}$ and cross-sampling $\hat{\rm X}$ as shown in Figure \ref{fig:model}:
\begin{equation}
\label{equ:self}
\hat{\rm M}=
\begin{cases}
    \bm{{\rm V}\!_M} {\rm softmax} (\bm{{\rm Q}_{M}}^\top \bm{{\rm K}_M}) \\
    \bm{{\rm V}\!_C} {\rm softmax} (\bm{{\rm Q}_{C}}^\top \bm{{\rm K}_C})
\end{cases}
\end{equation}
\begin{equation}
\label{equ:cross}
    \hat{\rm X}=\bm{{\rm V}\!_C} {\rm softmax} (\bm{{\rm Q}_{M}}^\top \bm{{\rm K}_C})
\end{equation}
where Eq.(\ref{equ:self}) and Eq.(\ref{equ:cross}) denote as the self-sampling attention and cross-sampling attention. Particularly, the upper formula of Eq.(\ref{equ:self}) calculates the self-sampling attention for multimodal knowledge $\rm M$, while the lower formula of Eq.(\ref{equ:self}) calculates the self-sampling attention for confounding factors $\rm C$. In the implementation, $\bm{{\rm Q}_M}$ and $\bm{{\rm Q}_C}$ are derived from $h({\rm X})$ and $f({\rm X})$. $\bm{{\rm K}_M}$ and $\bm{{\rm V}\!_M}$ are obtained from the current input sample, while $\bm{{\rm K}_C}$ and $\bm{{\rm V}\!_C}$ come from other samples in the training set and serve as global dictionary compressed from the whole training dataset. Specifically, we initialize this dictionary by using K-means clustering~\cite{hartigan1979algorithm} on all the embeddings of samples in the training set, such as region features.

\subsection{Implicit-aware Counterfactual Inference Block}
\label{sec:icfi}
In this study, we take advantage of the counterfactual technique~\cite{pearl2018book} and propose an \textbf{I}mplicit-aware \textbf{C}ounterfactual \textbf{I}nference (ICI) block to highlight the implicit relations beyond the explicit, thereby addressing the imbalance problem between them. Specifically, we treat the explicit relations as the direct effect in the counterfactual technique, and then reduce such direct effect to achieve the goal of reducing the importance of the explicit relations while highlighting the implicit relations. Therefore, there are also two questions to be answered: 1) how to analyze the direct effect of the explicit relations; 2) how to reduce such direct effect to improve the alignment of implicit phrase-region pairs. Next, we will answer the two questions, formulated as follows.

\textbf{Counterfactual Causal Graph} is leveraged to answer the question 1). As illustrated in Figure \ref{fig:model} (c), we formulate the causation between explicit and implicit relations through the path $\rm X \rightarrow \rm E \rightarrow \rm L$, which denotes the causal effect from the imbalanced explicit relations $\rm E$ to phrase-region locations $\rm L$. On this basis, we leverage counterfactual to analyze the causal effects. Following the counterfactual annotations in~\citet{pearl2018book} and \citet{NiuTZL0W21}, we denote ${\rm L}_{x,e}={\rm L}({\rm X}=x,{\rm E}=e)$ as the \emph{total effect} (TE). We block the path $\rm X \rightarrow \rm E$ to obtain the \emph{explicit direct effect} (EDE) on $\rm L$ as shown in Figure \ref{fig:model} (d), denoted as ${\rm L}_{x^*,e}={\rm L}({\rm X}=x^*,{\rm E}=e)$, which represents the value of $\rm L$ when we set $x$ to $x^*$. Note that only in the counterfactual world, $\rm X$ can be simultaneously set to different values $x$ and $x^*$. To reduce the EDE from TE, we aim to derive the \emph{explicit indirect effect} (EIE), denoted as follows:
\begin{equation}
\label{equ:EIE}
{\rm EIE} = {\rm TE} - {\rm EDE} = {\rm L}_{x,e} - {\rm L}_{x^*,e}
\end{equation}

\textbf{Implicit-aware Inference} is leveraged to answer the question 2). Upon obtaining the output representations $o_i$ from IDA block, we utilize ICI to reduce the direct effect of explicit relations in Eq.(\ref{equ:EIE}). In this context, the representations $o_i$ can be seen as value $x$ in ICI. For the value of $x^*$, we assume that the model will randomly guess with equal probability, denoted as follows:
\begin{equation}
\label{equ:none}
x=o_i, x^*=r
\end{equation}
where $r$ denotes a learnable parameter. Finally, we can compute the EIE representation (i.e., similarity matrix $\bm{{\rm A}_{n,m}}$) in Eq.(\ref{equ:EIE}), which are then employed to calculate similarities for the WPG task.  

\subsection{Weakly-supervised Optimization}
During the training stage, our access is limited to coarse-grained sentence-image pairs $(S,V)$, from which we derive the ground-truth label $y$ based on the matching relations of each sentence-image pair. Therefore, following \citet{ChenGN18}, we convert the phrase-region similarity matrix $\bm{{\rm A}_{n,m}}$ to sentence-image similarity matrix $\bm{{\rm A}_{T,T}}$ for weakly-supervised optimization. Specifically, we first compute the similarity value ${\rm sim}_i$ for the sentence-image pair $(S_i,V_i)$, i.e., ${\rm sim}_i = \frac{1}{n}\sum_{n}\max_{m}\bm{{\rm A}_{n,m}}$. On this basis, we then calculate the similarity values between the current sentence $S_i$ and all images $V$. Finally, we obtain the sentence-image similarity matrix $\bm{{\rm A}_{T,T}}$ encompassing all sentence-image pairs. After the argmax operation, we obtain the predicted label $\hat{y}$. To train our IECI approach end-to-end, we leverage the cross-entropy loss function, denoted as follows:
\begin{equation}
\label{equ:cp}
    \mathcal{L}_{wpg} = -\sum_{i=1}^T y_i\log \hat{y_i}
\end{equation}  

Besides, we incorporate a learnable parameter $\bm{r}$ in Eq.(\ref{equ:none}), which controls the sharpness of the distribution of $L(x^*,e)$. If $\bm{r}$ is inappropriate, EIE would be guided by TE or EDE. To avoid this, we leverage a Kullback-Leibler divergence to update $\bm{r}$ via back propagation:
\begin{equation}
\label{equ:kl}
\mathcal{L}_{kl} = \sum -p(y|x,e)\log \frac{p(y|x^*,e)}{p(y|x,e)} 
\end{equation}

Consequently, our optimization objective comprises both $\mathcal{L}_{wpg}$ and $\mathcal{L}_{kl}$, denoted as $\mathcal{L}_{total} = \mathcal{L}_{wpg} + \alpha \mathcal{L}_{kl}$, where $\alpha$ serves as a hyper-parameter responsible for balancing the loss between the WPG task and ICI block. 

\section{Experimental Settings}
\setlength{\tabcolsep}{1.8pt}
\begin{table*}[t]
\renewcommand{\arraystretch}{1.3}
\addtolength{\tabcolsep}{2.7pt}
\begin{center}
    \begin{small}
        \begin{tabular}{c|cccllc|cccccc}
\toprule[1.2pt]
\multicolumn{1}{l|}{\multirow{3}{*}{Approach}} & \multicolumn{6}{c|}{Flickr30K}                                                                                                                          & \multicolumn{6}{c}{COCO}                                                                                                                      \\ \cline{2-13} 
\multicolumn{1}{l|}{}                          & \multicolumn{2}{c|}{Implicit}                        & \multicolumn{2}{c|}{Explicit}                        & \multicolumn{2}{c|}{Full}                  & \multicolumn{2}{c|}{Implicit}                        & \multicolumn{2}{c|}{Explicit}                        & \multicolumn{2}{c}{Full}        \\ \cline{2-13} 
\multicolumn{1}{l|}{}                          & R@1            & \multicolumn{1}{c|}{R@5}            & R@1            & \multicolumn{1}{c|}{R@5}            & \multicolumn{1}{c}{R@1}   & R@5            & R@1            & \multicolumn{1}{c|}{R@5}            & R@1            & \multicolumn{1}{c|}{R@5}            & R@1            & R@5            \\ \hline
KAC-Net                                        & 31.99          & \multicolumn{1}{c|}{51.14}          & 49.11          & \multicolumn{1}{l|}{63.50}          & 38.71$^\sharp$                     & -              & 39.51          & \multicolumn{1}{c|}{56.85}          & 54.81          & \multicolumn{1}{c|}{73.72}          & 45.88          & 66.27          \\
ARN                                            & 33.81          & \multicolumn{1}{c|}{49.87}          & 46.13          & \multicolumn{1}{l|}{64.50}          & 34.87                     & 50.42          & 40.35          & \multicolumn{1}{c|}{57.68}          & 57.07          & \multicolumn{1}{c|}{74.98}          & 41.93          & 58.27          \\
KPRN                                           & 30.95          & \multicolumn{1}{c|}{44.50}          & 47.31          & \multicolumn{1}{l|}{62.93}          & 33.41                     & 47.33          & 35.31          & \multicolumn{1}{c|}{55.22}          & 55.77          & \multicolumn{1}{c|}{76.51}          & 38.30          & 57.00          \\
InfoGround                                     & 44.72          & \multicolumn{1}{c|}{74.79}          & 55.07          & \multicolumn{1}{c|}{80.87}          & \multicolumn{1}{c}{47.88$^\dagger$} & 76.63$^\dagger$          & 45.66          & \multicolumn{1}{c|}{74.08}          & 61.17          & \multicolumn{1}{c|}{84.67}          & 51.67$^\dagger$          & 77.69$^\dagger$          \\
ALBEF                                          & 56.40          & \multicolumn{1}{c|}{78.27}          & 69.37          & \multicolumn{1}{c|}{85.03}          & \multicolumn{1}{c}{57.64} & 77.56          & 52.00          & \multicolumn{1}{c|}{76.23}          & 66.19          & \multicolumn{1}{c|}{84.06}          & 54.22          & 76.34          \\
CL\&KD                                         & 50.33          & \multicolumn{1}{c|}{73.75}          & 62.50          & \multicolumn{1}{l|}{82.00}          & 53.10$^\natural$                     & -              & 50.37          & \multicolumn{1}{c|}{75.02}          & 64.42          & \multicolumn{1}{c|}{83.78}          & 51.36          & 74.98          \\
ReIR                                           & 57.98          & \multicolumn{1}{c|}{\textbf{78.72}}          & 69.65          & \multicolumn{1}{l|}{85.33}          & 59.27$^\S$                     & -              & 54.01          & \multicolumn{1}{c|}{77.60}          & 66.77          & \multicolumn{1}{c|}{84.45}          & 55.26          & 76.72          \\
BLIP                                           & 20.31          & \multicolumn{1}{c|}{41.51}          & 26.62          & \multicolumn{1}{l|}{62.62}          & 23.30                     & 57.07              & 26.99          & \multicolumn{1}{c|}{63.19}          & 34.14          & \multicolumn{1}{c|}{70.02}          & 31.15          & 67.13          \\\hline
\textbf{IECI}                                   & \textbf{61.32} & \multicolumn{1}{c|}{78.36} & \textbf{72.37} & \multicolumn{1}{l|}{\textbf{86.27}} & \textbf{62.29}            & \textbf{79.28} & \textbf{56.32}          & \multicolumn{1}{c|}{\textbf{78.01}} & \textbf{68.62}          & \multicolumn{1}{c|}{\textbf{85.25}} & \textbf{56.92}          & \textbf{78.31} \\
w/o IDA                                       & 57.72          & \multicolumn{1}{c|}{77.16}          & 71.10          & \multicolumn{1}{l|}{84.96}          & 59.07                     & 77.87          & 53.96          & \multicolumn{1}{c|}{76.88}          & 67.19 & \multicolumn{1}{c|}{84.13}          & 54.92 & 77.25          \\
w/o ICI                                       & 58.09          & \multicolumn{1}{c|}{77.79}          & 71.87          & \multicolumn{1}{l|}{85.33}          & 59.50                     & 78.51          & 54.44 & \multicolumn{1}{c|}{77.12}          & 67.61          & \multicolumn{1}{c|}{84.71}          & 55.05          & 77.17          \\
w/o Both                                       & 56.05          & \multicolumn{1}{c|}{76.48}          & 69.32          & \multicolumn{1}{l|}{83.89}          & 57.87                     & 77.78          & 52.20          & \multicolumn{1}{c|}{76.07}          & 66.05          & \multicolumn{1}{c|}{83.57}          & 54.17          & 76.54 \\ 
\bottomrule[1.2pt]
\end{tabular}
\setlength{\belowcaptionskip}{-2.0 ex}
\caption{Comparison of several state-of-the-art baselines and our approaches on both the Flickr30K and COCO datasets (training sets). \emph{Implicit}, \emph{Explicit} and \emph{Full} represent the evaluation results on our annotated implicit and explicit datasets, the original Flickr30K-Entities dataset, respectively. The result with symbol $\sharp$ is retrieved from \citet{ChenGN18}; those with $\dagger$ are from \citet{GuptaVC0KH20}; this with $\natural$ is from \citet{WangHLXY021} and this with $\S$ is from \citet{LiuWM021}. The symbol - denotes the results are not reported by these papers.}
\label{tab:results}
    \end{small}
\end{center}
\end{table*}

\subsection{Data Annotation and Settings} 
\label{sec:datasets}
To evaluate the effectiveness of our IECI approach in WPG, we construct an implicit-enhanced dataset based on the Flickr30K-Entities~\cite{PlummerWCCHL15} dataset. Specifically, for the annotation of the implicit-enhanced dataset, we first summarize four main types of implicit relations through preliminary annotation, and estimate their ratios by analyzing 100 randomly-selected annotated samples within Flickr30K-Entities. Specifically, the four types of implicit relations (as illustrated in Figure \ref{fig:annotation}) are introduced as follows:
\begin{itemize}
\setlength{\itemsep}{0pt}
\setlength{\parsep}{0pt}
\setlength{\parskip}{0pt}
    \item \textbf{CU} denotes the model needs extra knowledge to understand commonsense, e.g., ``\emph{support}'' needs to understand the commonsense of \emph{holding a flag} to precisely predict the region.
    \item \textbf{CCU} denotes the model needs the context of sentences to understand the meaning of phrases, e.g., ``\emph{another person}'' can hardly correspond to the exact region without context.
    \item \textbf{SRU} represents the position relation between two target objects in the sentence, e.g., the region of ``\emph{next to another man}'' contains the position relation between \emph{two men}.
    \item \textbf{NU} represents a phrase that may correspond to multiple regions, e.g., ``\emph{three of them}'' corresponds to three regions in the image.
\end{itemize}
For annotation, we assign two annotators to tag each phrase-region pair and the \emph{Kappa} consistency check value of the annotation is 0.85. When two annotators cannot reach an agreement, an expert will make the final decision, ensuring the quality of data annotations. Furthermore, we annotate 2K coarse-grained sentence-image samples, including 15K fine-grained phrase-region pairs in the original Flickr30K-Entities dataset. After the manual annotation, we finally obtain 1.4K implicit phrase-region pairs. The ratio between the implicit and explicit phrase-region pairs is about 1:9, which shows an extreme imbalance between them, thereby encouraging us to address the challenge of highlighting the implicit relations beyond the explicit. 

For training WPG, following \citet{GuptaVC0KH20}, we leverage the training sets from Flickr30K and COCO datasets. For inference, we maintain the same setting of validation and test splits as the original Flickr30K-Entities (\emph{Full}) dataset, and obtain the \emph{Implicit}, \emph{Explicit} datasets based on these splits.

\subsection{Baselines}
We choose several state-of-the-art baselines for WPG to compare performance with our IECI approach, described as follows.
\begin{itemize}[left=0pt]
\item \textbf{KAC-Net}~\cite{ChenGN18} explores the consistency in visual and language as complementary external knowledge. 
\item \textbf{ARN}~\cite{Liu0WZMH19} builds the correspondence between image region and query in an adaptive manner. 

\item \textbf{KPRN}~\cite{LiuLWZSH19} models the relation between target and contextual entities. 
\item \textbf{InfoGround}~\cite{GuptaVC0KH20} leverages contrastive learning to maximize a lower bound between region features and contextualized word representations.

\item \textbf{ALBEF}~\cite{LiSGJXH21} introduces a contrastive loss to align the image and text representations before fusing them into cross-modal attention.

\item \textbf{CL\&KD}~\cite{WangHLXY021} focuses on distilling knowledge from a generic object detector under the framework of contrastive learning.

\item \textbf{ReIR}~\cite{LiuWM021} incorporates coarse-to-fine object refinement and entity relation modeling into a two-stage deep network.

\end{itemize}
    Moreover, an advanced multimodal pre-training model \textbf{BLIP}~\cite{0001LXH22} is leveraged to compare with IECI. Here, it should be noted that BLIP can not directly fine-tune the WPG task, since BLIP is pre-trained with the coarse-grained sentence-image pairs. Thus, we treat the phrase-region pairs as the coarse-grained pairs to fine-tune the WPG task. 

\subsection{Implementation Details and Metrics}
In our experiments, we leverage open-source codes to obtain experimental results of all the baselines on both \emph{Implicit} and \emph{Explicit} datasets, and re-implement the results of several baselines by their open-source codes on \emph{Full} datasets. The hyper-parameters of these baselines reported by their public papers maintain the same setting, and the others are tuned according to the validation set. Specifically, for phrase encoder, the parameters of BERT are following \citet{DevlinCLT19}. For region encoder, we utilize a pre-trained Faster R-CNN on the Visual Genome~\cite{KrishnaZGJHKCKL17} dataset to generate a maximum number of 100 object regions. For IECI, we employ the Adam optimizer with an initial learning rate of 1e-5 for training. The regularization weight of parameters is 1e-4. The hyper-parameter $\alpha$ is set to be 0.1 and the batch size is set to be 64. The layers of the self-sampling attention and cross-sampling attention in IDA are both set to be 6. The overall training parameters of our approach are 0.16B. All the baselines are reproduced using PyTorch on a machine equipped with an NVIDIA GeForce RTX 3090, an Intel(R) Xeon(R) E5-2650 v4 CPU (2.20 GHz), CUDA version 11.7, and the PyTorch 1.7.1 library with Python 3.6.13, running on Ubuntu 20.04.1 LTS. To facilitate the corresponding research in this direction, all codes alongside our implicit-enhanced dataset will be released via Github \footnote{Github link will come soon.}.

Besides, the WPG performance is evaluated using Recall@$k$, $k \in (1,5)$, which measures the fraction of phrases  for which the ground-truth bounding box exhibits an IoU $\geqslant$ 0.5 with any of the top-$k$ predicted boxes. Moreover, $t$-test\footnote{https://docs.scipy.org/doc/scipy/reference/stats.html} is used to evaluate the significance of the performance difference.

\section{Results and Discussions}
\label{sec:analysis}

\subsection{Experimental Results}
Table \ref{tab:results} shows the performance comparison of different approaches. From this table, we can see that: \textbf{1)} The performances of all approaches on the \emph{Implicit} dataset are consistently lower than the \emph{Explicit} dataset. This indicates that modeling the implicit relations is more challenging than the explicit, and encourages us to consider such implicit relations problem in WPG. Besides, we also observe that the performance of both \emph{Implicit} and \emph{Explicit} datasets is improved, and we analyze the reason is that there may be some overlap between the implicit and explicit relations in the current dataset, leading to performance improvements in explicit relations when employing causal inference. \textbf{2)} The performance improvements of our IECI approach on the \emph{Implicit} dataset are larger than those on the \emph{Explicit} dataset. This indicates that IECI can effectively address the challenge of highlighting the implicit relations beyond the explicit. Furthermore, our IECI approach consistently performs better than all other approaches. Impressively, compared to the best-performing ReIR approach, our IECI approach achieves the average R@1 improvements of 2.83\%, 2.29\% and 2.34\% on all three \emph{Implicit}, {Explicit} and \emph{Full} datasets, respectively. Significance test shows that all these improvements are significant ($p$-value < 0.01). These demonstrate that IECI can better highlight the implicit relations and meantime maintain the performance of the explicit. Particularly, we observe that the performance of BLIP is quite worse than IECI. This is reasonable since BLIP is pre-trained on the coarse-grained sentence-images pairs, thus is not suitable for aligning the fine-grained phrase-region pairs. \textbf{3)} Our IECI approach trained on both the Flickr30K and COCO datasets still outperforms all other approaches. This justifies the robustness of IECI, and again encourages us to consider the important implicit relations in the WPG task.

\begin{figure}
\begin{center}
\subfloat{\includegraphics[scale=0.40]{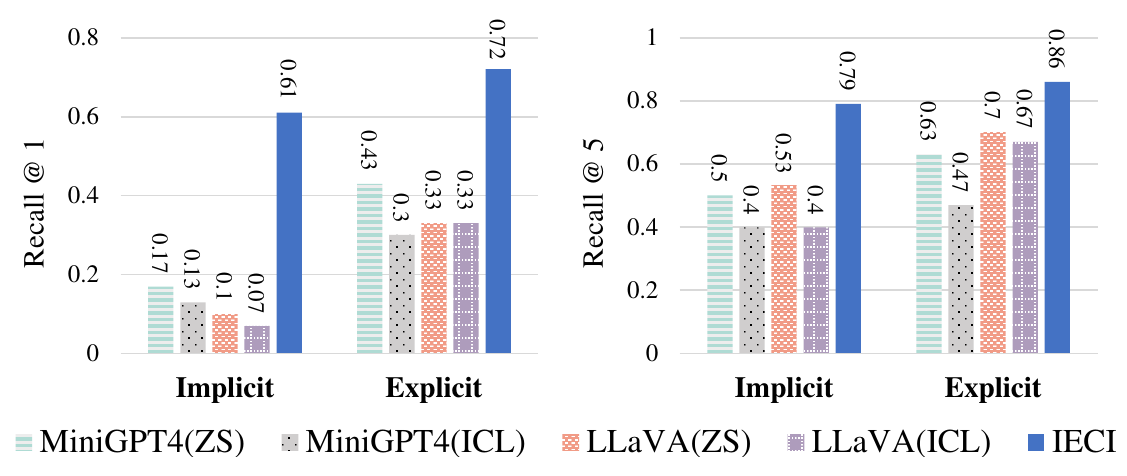}}
\setlength{\belowcaptionskip}{-4.0 ex}
\caption{Comparison performance between multimodal LLMs (i.e., MiniGPT4-13B, LLaVA-13B) and our IECI approach, where ZS and ICL represent zero-shot and in-context learning evaluation methods for LLMs.}
\label{fig:llm}
\end{center}
\end{figure}

\subsection{Contributions of Causal Inference}
To further investigate the influence of key causal components within our IECI approach, we conduct a series of ablation studies as shown in Table \ref{tab:results}. From this table, we can see that: \textbf{1) w/o IDA} exhibits inferior performance compared to IECI, with an average decrease in R@1 by 2.98\% ($p$-value < 0.01) and 1.35\% ($p$-value < 0.05) on the \emph{Implicit} and \emph{Explicit} datasets, respectively. This further justifies that our proposed IDA block can effectively model the implicit relations, encouraging us to leverage the intervention technique to mitigate the confounding bias. \textbf{2) w/o ICI} also shows inferior performance compared to IECI, with an average R@1 decrease of 2.56\% ($p$-value < 0.01) and 0.76\% ($p$-value < 0.05) on the \emph{Implicit} and \emph{Explicit} datasets, respectively. This further justifies that ICI block can effectively highlight the implicit relations beyond the explicit, encouraging us to leverage the counterfactual technique to address the imbalance problem. \textbf{3) w/o Both} yields a significant average decrease in R@1 by 4.70\% ($p$-value < 0.01) on the \emph{Implicit} dataset, which excludes both IDA and ICI blocks. This again justifies the effectiveness of our IECI approach in modeling and highlighting the implicit phrase-region matching relations.

\subsection{Comparison with Multimodal LLMs}
\label{sec:llms}
Recently, the emergence of large language models (LLMs) has demonstrated their remarkable abilities across various fields and tasks. Since GPT-4 is closed-sourced and currently cannot evaluate the multimodal tasks, we choose two advanced open-sourced multimodal LLMs, i.e., MiniGPT4-13B~\cite{abs-2304-10592} and LLaVA-13B~\cite{abs-2304-08485}, to verify the effectiveness of IECI. Specifically, we first randomly select 30 sentence-image examples with the implicit and explicit relations for comparisons on the WPG task. Then, we select two well-studied evaluation methods for LLMs as follows: \textbf{1) Zero-Shot (ZS).} Following the ZS setting for evaluating LLMs proposed by~\citet{abs-2302-04023}, given a prompt including the task instructions, sentence-image pair and the bounding box list of each phrase, we utilize multimodal LLMs to ``\emph{generate the bounding boxes from the list that can reflect the given phrase}''. \textbf{2) In-Context Learning (ICL).} Following the ICL setting for evaluating LLMs proposed by~\citet{abs-2302-04931}, building upon the ZS, we introduce one additional demonstration example ``\emph{phrase, boxes;...;phrase, boxes}'' as the prompt. As shown in Figure \ref{fig:llm}, we observe that IECI outperforms the multimodal LLMs by a large margin, indicating that the multimodal LLMs still face challenges in understanding the deep multimodal semantics. Moreover, we find that the performance of ZS is better than ICL, since the image encoders BLIP-2 and CLIP within MiniGPT4-13B and LLaVA-13B lack the ability of ICL as reported in \citet{abs-2301-12597} and \citet{RadfordKHRGASAM21}.

\begin{figure}
\begin{center}
\subfloat{\includegraphics[scale=0.44]{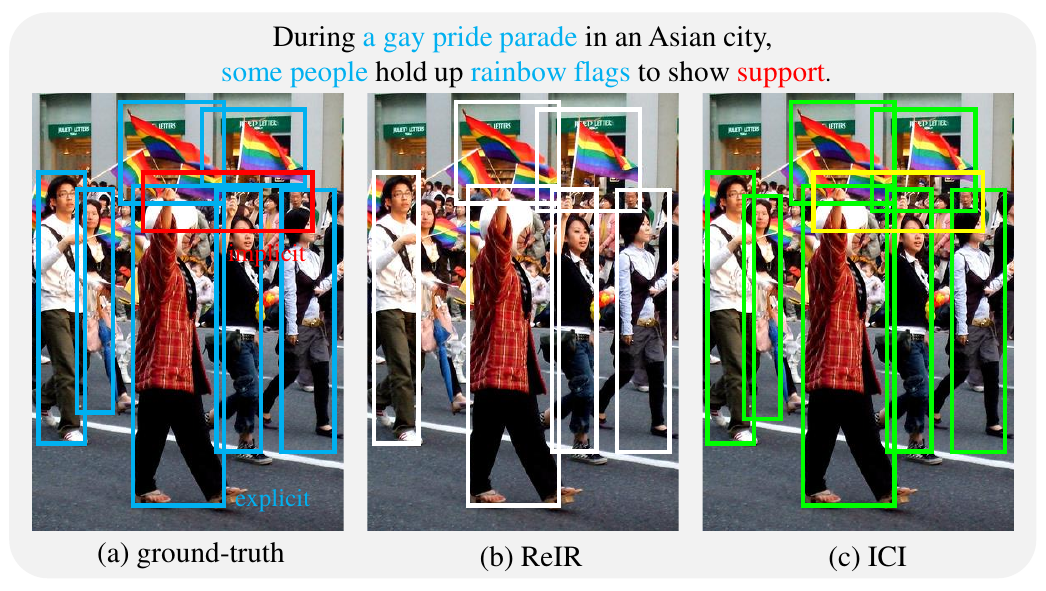}}
\setlength{\belowcaptionskip}{-4.0 ex}
\caption{A sentence-image example from our implicit-enhanced dataset, along with their ground-truth phrase-region pairs (a), predicted regions by best-performing baseline ReIR (b), and predicted regions by our IECI approach (c).}
\label{fig:case}
\end{center}
\end{figure}

\subsection{Visualization Study}
We conduct a visualization analysis of IECI in our implicit-enhanced dataset. As illustrated in Figure \ref{fig:case}, we can see that: \textbf{1)} The prediction of the region corresponding to the phrase ``\emph{support}'' (red box in Figure \ref{fig:case} (a)) is challenging, which requires contextual and commonsense knowledge. This further indicates the difficulty in precisely aligning such implicit phrase-region pairs. \textbf{2)} We compare the performance of the best-performing baseline ReIR and IECI, as shown in Figure \ref{fig:case} (b) and (c). For the phrase ``\emph{support}'', ReIR fails to predict the corresponding region, while IECI successfully identifies the correct region (yellow box). This again justifies the effectiveness of IECI in precisely identifying such implicit phrase-region matching.

\subsection{What Would We Do Next?}
Our proposed IECI approach has demonstrated promising results in comprehensive experiments, showcasing its effectiveness in addressing the challenges of modeling the implicit relations and highlighting them beyond the explicit. However, we believe that there are still three potential directions to precisely predict implicit phrase-region matching relations. Despite this is not the focus of this paper, we believe that these three directions can not only further improve the performance of the WPG task, but also facilitate related research in this direction, as described below.

Firstly, for \textbf{Multimodal Representation}, our IECI approach has limitations in effectively representing underlying features. However, the existing multimodal pre-training models (e.g., CLIP~\cite{RadfordKHRGASAM21}, BLIP~\cite{0001LXH22}) are mostly pre-trained on coarse-grained sentence-image pairs, making them lack the ability of fine-grained multimodal semantics understanding, as reported in Section \ref{sec:llms}. In our future work, we would like to incorporate the multimodal LLMs (e.g., MiniGPT4~\cite{abs-2304-10592}, LLaVA~\cite{abs-2304-08485}) to enhance the multimodal representation abilities of our approach for the WPG task.

Secondly, for \textbf{Knowledge Injection}, we observe that the ratio of commonsense understanding is 34.5\% as illustrated in Figure \ref{fig:annotation}, which further inspires us to consider integrating the external knowledge (e.g., multimodal knowledge graph~\cite{abs-2202-05786}) to assist in capturing the implicit relations.

Finally, for \textbf{Evaluation of Each Implicit Relation}, due to the imbalanced proportion of implicit relations in the whole dataset, the sample scale of each implicit relation is relatively small (see Section \ref{sec:datasets}), making it difficult to comprehensively evaluate the effectiveness of our IECI approach on each implicit relation. In our future work, we would like to leverage the multimodal LLMs to automatically annotate (e.g., the automatic annotation approach proposed by~\citet{abs-2306-04349}) different types of implicit relations, which may promote the further research in this direction.

\section{Conclusion}  
In this paper, we propose an \textbf{I}mplicit-\textbf{E}nhanced \textbf{C}ausal \textbf{I}nference (IECI) approach to address the implicit phrase-region matching relations problem inside WPG. The key idea of IECI is to utilize the causal inference (i.e., intervention and counterfactual techniques) to effectively model the implicit relations and highlight them beyond the explicit. To comprehensively evaluate IECI, we construct a specialized implicit-enhanced dataset. Detailed experimental results on this dataset demonstrate the superior performance of IECI over several state-of-the-art baselines. 

In our future work, we would like to introduce more information (e.g., multimodal knowledge graph~\cite{abs-2202-05786}) to assist in aligning implicit phrase-region pairs. Moreover, we would like to transfer IECI to other tasks which also have the implicit relation problems, such as Referring Expression Comprehension and video grounding.

\section*{Ethics Statement}
\textbf{Data Disclaimer.}
We construct a high-quality implicit-enhanced dataset based on Flickr30K-Entities, a dataset which is widely used by other academics and are typically accessible to the public. Therefore, our proposed dataset does not involve any sensitive information that may harm others. 

\noindent\textbf{Human Annotation.} 
When recruiting annotators for this study, we claim that all potential annotators are free to choose whether they want to participate, and they can withdraw from the study anytime without any negative repercussions. Thus, the establishment of our dataset is compliant with ethics.

\section*{Acknowledgments}
We thank our anonymous reviewers for their helpful comments. This work was supported by three NSFC grants, i.e., No.62006166, No.62076175 and No.62076176. This work was also supported by a Project Funded by the Priority Academic Program Development of Jiangsu Higher Education Institutions (PAPD).

\section*{References}
\bibliographystyle{lrec-coling2024-natbib}
\bibliography{lrec-coling2024-example}

\bibliographystylelanguageresource{lrec-coling2024-natbib}
\bibliographylanguageresource{languageresource}

\end{document}